\RequirePackage{fix-cm}
\documentclass[smallextended]{svjour3}       % onecolumn (second format)
\smartqed  % flush right qed marks, e.g. at end of proof
\usepackage{graphicx}
\usepackage{ragged2e}
\usepackage{mathptmx}       % selects Times Roman as basic font
\usepackage{helvet}         % selects Helvetica as sans-serif font
\usepackage{courier}        % selects Courier as typewriter font
\usepackage{type1cm}
\usepackage{amssymb}
\usepackage{url}
\usepackage{hyperref}
\usepackage{footnote}
\usepackage[linesnumbered,ruled,vlined]{algorithm2e}
\usepackage{graphicx}
\usepackage{array}
\usepackage{tabu}
\usepackage[justification=centering]{caption}
\usepackage{float}
\usepackage{mathtools}
\usepackage{amsmath}
\usepackage{cite}
\usepackage{comment}
\usepackage{dirtytalk}
\usepackage{algpseudocode}
\usepackage[section]{placeins}
\usepackage{multirow,multicol}
\usepackage{chngcntr}   % activate if the above 3 fonts are
\usepackage{makeidx}         % allows index generation
\usepackage{graphicx}        % standard LaTeX graphics tool
\usepackage{multicol}        % used for the two-column index
\usepackage[bottom]{footmisc}% places footnotes at page bottom
\begin{document}

\title{Efficient Feature Selection techniques for Sentiment Analysis%\thanks{Grants or other notes
}

\author{Avinash M*        \and
	Sivasankar E %etc.
}


\institute{
	*corresponding author\\
	Avinash M \at
	Software Engineer,Samsung R and D Institute India, Bengaluru,Bagmane Constellation Business Park, Outer Ring Road, Doddanekundi Circle, Marathahalli Post, Bengaluru, Karnataka 560037 \\
	\email{avinash.sai001@gmail.com}           %  \\
	\and
	Sivasankar E \at
	Assistant Professor,Department of Computer Science,National Institute of Technology, Tiruchirappalli,
	Tanjore Main Road, National Highway 67, Near BHEL Trichy, Tiruchirappalli, Tamil Nadu 620015 \\
	\email{sivasankar@nitt.edu} 
}

\date{Received: date / Accepted: date}

\maketitle

\begin{abstract}
Sentiment analysis is a domain of study that focuses on identifying and classifying the ideas expressed in the form of text into positive, negative and neutral polarities. Feature selection is a crucial process in machine learning. In this paper, we aim to study the performance of different feature selection techniques for sentiment analysis. Term Frequency Inverse Document Frequency (TF-IDF) is used as the feature extraction technique for creating feature vocabulary. Various Feature Selection (FS) techniques are experimented to select the best set of features from feature vocabulary. The selected features are trained using different machine learning classifiers Logistic Regression (LR), Support Vector Machines (SVM), Decision Tree (DT) and Naive Bayes (NB). Ensemble techniques Bagging and Random Subspace are applied on classifiers to enhance the performance on sentiment analysis. We show that, when the best FS techniques are trained using ensemble methods achieve remarkable results on sentiment analysis. We also compare the performance of FS methods trained using Bagging, Random Subspace with varied neural network architectures. We show that FS techniques trained using ensemble classifiers outperform neural networks requiring significantly less training time and  parameters thereby eliminating the need for extensive hyper-parameter tuning. 

\end{abstract}
\section{Introduction}
Feature engineering is the process of constructing feature set using the properties of data that leverage the performance of machine learning algorithms. However, these features can be of very high dimensions and are difficult to train. Dimensionality reduction is one of the most popular methods used for mapping higher dimensional features to lower dimension meaningful representations \cite{Maaten:Postma}. Feature Extraction and Feature Selection are the two types of dimensionality reduction techniques \cite{Tang:Alelyani}. Feature extraction constructs new features using combinations of original features and projects them into a lower dimensional space. On the other hand, Feature Selection (FS) aims to select a subset of features that are highly relevant using a criterion measure. Feature Selection techniques can be broadly categorized into supervised and unsupervised. Filter and Wrapper methods are some of the Supervised FS techniques. Wrapper models form clusters of feature subsets from the entire feature set. They heuristically select features by measuring the accuracy of machine learning algorithm trained on the selected feature subset \cite{Das}. However, Wrapper methods are computationally expensive and infeasible if there are huge number of features. Wrappers methods may try to overfit with small training data and more parameters \cite{Das}. Filter methods on the other hand, select features using a statistical measure, for example correlation. Relief  \cite{Robnik:Koneneko}, Chi-Square \cite{Plackett} and Information Gain are some of the filter methods. Filter methods are independent of the training algorithm and hence prevent feature bias from interacting with training algorithm bias. In this paper, we concentrate on Filter methods for sentiment analysis.       

The rapid advancements of technology has led to the invention of social media applications like facebook, twitter and e-commerce companies  Amazon, Alibaba. Sentiment analysis plays a significant role in many real world applications. The aim of sentiment analysis is to classify text sentences into either of the predefined polarities positive, negative and neutral. In early days, sentiment analysis was performed using lexicon based methods.
Bag-of-Words (BoW) \cite{Harris} and Term Frequency-Inverse Document Frequency (TF-IDF) have been the most popular feature extraction techniques. Machine learning classifiers Support Vector Machines (SVM) and Naive Bayes (NB) were used to train these features \cite{Pang:Lee:Vaithyanathan}. A variant of Naive Bayes and Support Vector Machines, NB-SVM achieved excellent results on text classification \cite{Wang:Manning}. With the growth of deep learning, several neural network architectures were proposed to deal with text data. Recurrent Neural Networks (RNN) were introduced for handling sequential information. However, RNN suffered from vanishing \cite{Hochreiter:Bengio} and exploding gradient problems \cite{Pascanu:Mikolov:Bengio}. Hence, Long Short Term Memory Network (LSTM) was introduced \cite{Hochreiter:Schmidhuber}. Convolution Neural Network (CNN) architectures were proposed for sentence classification \cite{Kim} and text classification \cite{Conneau:Lecun}. However, all these architectures require extensive hyperparameter tuning and large number of parameters for training. 

Often, training a single machine learning classifier might not be able to draw robust decision boundary for classification. The classifier might show high bias leading to overfitting or high variance leading to underfitting. This can be avoided by training multiple classifiers together and then use their combined predictions. This is called Ensemble Learning. Bagging \cite{Breiman} and Random Subspace \cite{Barandiaran} are the two most popular ensemble methods. Ensemble techniques have shown to perform better than individual classifiers. 

The objective of this paper is to perform extensive evaluation of different FS techniques for sentiment analysis. The features selected are trained using different machine learning classifiers Logistic Regression (LR), Support Vector Machines (SVM), Decision Trees (DT) and Naive Bayes (NB). To further enhance the performance, ensemble techniques Bagging and Random Subspace are used to train classifiers. We show that FS techniques trained using ensemble techniques achieve remarkable performance on sentiment analysis. We also compare the performance of best FS techniques trained using ensemble methods with neural network architectures on standard benchmark datasets. We show that FS techniques trained using Bagging and Random Subspace achieved better results compared to neural networks with significantly less time and model parameters thereby eliminating the need for extensive hyper-parameter tuning.

The main contributions of the paper include:
\begin{itemize}
  \item Different Feature Selection techniques are experimented for sentiment analysis.
  \item The features are trained using machine learning classifiers and ensemble techniques, bagging and random subspace.
 \item Effectiveness of FS techniques trained using Ensemble classifiers is experimented on several standard benchmark datasets and compared with various neural network architectures.
\end{itemize}
\section{Related Work}
Morinaga \cite{Morinaga} performed mining product reputations on web using characteristic word extraction, word co-occurrence analysis. This served as a starting point for opinion mining on customer reviews. Machine learning classifiers like SVM, Naive Bayes were trained to perform sentiment classification on product reviews which outperformed human crafted features  \cite{Pang:Lee:Vaithyanathan}. Bag-of-Words was the feature extraction technique used for composing feature vocabulary. However, selecting words based on the count threshold often ignores the most important words which occur rarely. Hence TF-IDF was introduced which proved to be better feature extraction technique \cite{Jones}.

Although, a feature vocabulary is formed, it is important to reduce the feature dimensions by selecting the features that are vital in expressing the sentiment. Therefore, FS techniques are crucial in reducing the training time and for improving the performance of training algorithm \cite{Tim:Irena}. Information Gain (IG), Mutual Information (MI), Chi-Square were used as the FS techniques for performing sentiment analysis on chinese documents \cite{Tan:Zhang}. Entropy Weighted Genetic Algorithm (EWGA), a combination of IG and genetic algorithm improved the accuracy of sentiment classification \cite{Abbasi:Chen}. Fisher's discriminant ratio is used as the feature selection criterion for performing sentiment classification.
\cite{Wang:Li:Wei}.

A variant of Categorical Proportion Difference (CPD), Categorical Probability Proportion Difference (CPPD) is proposed to perform sentiment classification \cite{Agarwal:Mittal:Acl}. They also used Parts-Of-Speech (POS) based adjectives and adverbs as the features in conjunction with features extracted using IG and Minimum Redundancy Maximum Relevancy (mRMR). These features are used to perform sentiment analysis on 4 different datasets (Books, DVD, Electronics, Kitchen) \cite{Agarwal:Mittal}. Chakrit \cite{Pong:Kaewmak} improved the performance of sentiment analysis for teacher evaluation using feature selection techniques and ensemble technique Voting. They used Chi-Square as the feature selection technique and used ensemble techniques to improve the classifiers performance. An adaptive two stage feature selection is proposed for sentiment classification \cite{Chi:Siew}. In the first stage feature scores are generated from training data and then weights the scores based on the test data samples.

An improved Chi-Square FS is proposed based on Chi-Square for performing Arabic Text Classification \cite{Bahassine}. Oussous \cite{Oussous} performed a study on the effects of pre-processing and ensemble learning for Arabic text classification.  An evolutionary ensemble learning is proposed for reducing the problem of domain adaptation of sentiment analysis \cite{Lopez}. Memetic FS technique is proposed based on Frequency difference for multi-label text categorization \cite{Lee:Yu}. Multivariate filter method Multivariate Relative Discriminative Criterion (MRDC) is proposed based on the Relative discriminative criterion (RDC) and correlation between features \cite{Labani}.

Recently, Sentiment Classification is performed using feature fusion. A rich feature set is formed by the fusion of word embedding, sentiment, statistical and linguistic knowledge \cite{Abdi}. An Adaptive Fusion and Category-Level Dictionary Learning Model is proposed for Multi-View Human Action Recognition \cite{Gao:Xuan}. An adaptive fusion of information from multiple domains is obtained using feature level fusion and score level fusion. A Cognitive-inspired class-statistics matching method with triple-constraint (CSTC) is introduced for 3D object retrieval \cite{Wang:Wan}. For each feature in the class, Gaussian Probabilistic models are built which captures generality.

In this paper, an extensive performance analysis is conducted on various FS techniques for performing sentiment classification.
\section{Methodology}
\begin{figure}
	\centering
	\includegraphics[width=200mm]{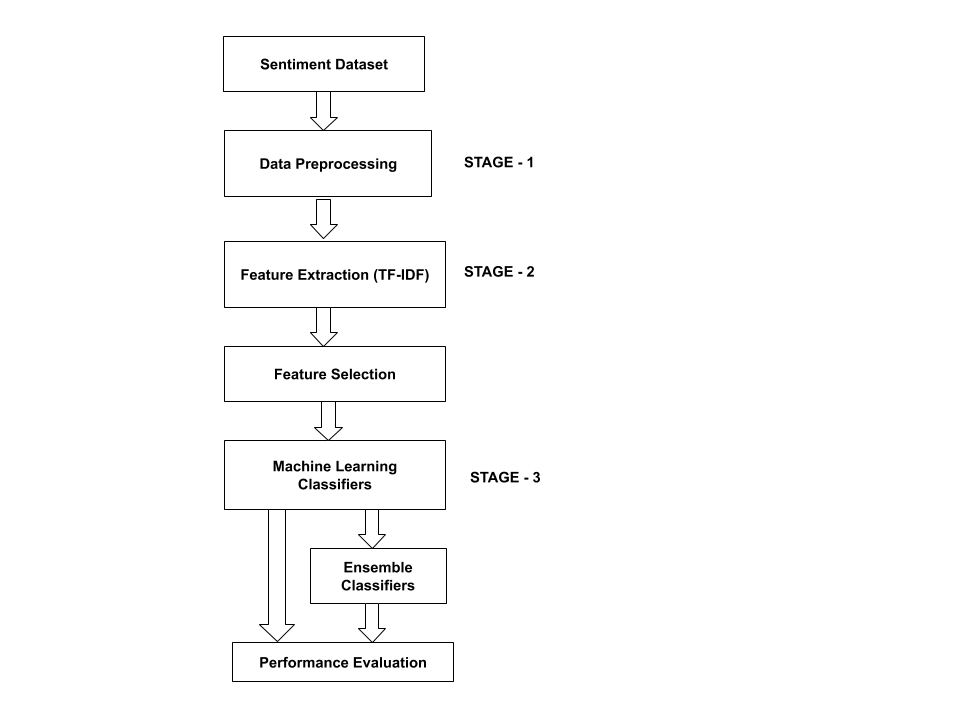}
	\caption{Flow diagram of the proposed analysis}
	\label{fig:1}       % Give a unique label
\end{figure}

The proposed analysis is performed in 4 stages. Preprocessing is done in first stage which involves cleaning the text data. Feature extraction and Feature selection is performed in stage 2. Stage 3 involves training the features using machine learning classifiers. Ensemble techniques are applied on classifiers and performance is evaluated in the final stage. Figure \ref{fig:1}
describes the stages involved in the experimentation.
\subsection{Preprocessing}
The reviews are preprocessed as follows:
\begin{itemize}
    \item The entire text is lower cased.
    \item Any character not in \textit{a-zA-Z} is filtered out.
    \item Stopwords are not removed.
    \item Words like \textit{can't} are replaced by cannot.
\end{itemize}

\subsection{TF-IDF}
TF-IDF is an unsupervised feature extraction algorithm which deals at the lexical level of data. It is based upon the frequency of the words in the text.
TF stands for term frequency which calculates how many times that word has occurred in the documents. It is calculated as:
\begin{equation}
TF(w)=\frac{\text{Number of times word $w$ appears in a document $d$}}{\text{Total number words in a document $d$}}\
\end{equation}
Inverse document frequency gives more importance to the rarely occurring words in the documents. It is calculated as:
\begin{equation}
IDF(w) = log_e \frac{\text{Total number of documents }}{\text{Total number of documents with word $w$ in them}}
\end{equation}  
TF-IDF score is calculated as:
\begin{equation}
TF-IDF(w) =  TF(w) \times IDF(w)
\end{equation}
\subsection{Feature Selection Techniques}
We use the following notations for all the feature selection techniques:
$n_{f}$ denotes frequency of documents containing feature $f$. $n_{\overline{f}}$ denotes frequency of documents not containing feature $f$. $n_{c_{k}}$ denotes the frequency of documents belonging to category $c_{k}$. $n_{c_{\overline{k}}}$  denotes the frequency of documents not belonging to category $c_{k}$. $n_{f,c_{k}}$ denotes the frequency of the feature $f$ for class $c_{k}$. $n_{f,c_{\overline{k}}}$ denotes the frequency of the documents containing feature $f$ but not in class $c_{k}$. $n_{\overline{f},c_{k}}$ denotes frequency of the documents that doesn't contain feature $f$ belonging to class $c_{k}$. $n_{\overline{f},c_{\overline{k}}}$ denotes the frequency of documents that doesn't contain feature $f$ that doesn't belong to category $c_{k}$. $N$ is the number of documents.

\subsubsection{Odds Ratio (OR)}
Odds Ratio measures the odds of a feature occurring in positive category. When the feature occurs more in positive category then it has a higher score. If the feature occurs more in negative category then its has a lesser score. If the feature occurs equally in both the categories then its score is 0. It is calculated as :
\begin{equation}
OR(f) = log_e \frac{OR(f,c_{1})}{OR(f,c_{0})}
\end{equation}
\subsubsection{Chi-Square (Chi)}
Chi-Square measures the relationship between variables. If the value is zero means that the variables are completely independent. If the variables have higher ${\chi}^2$, they are very closely related to each other. In sentiment analysis, ${\chi}^2$ test is used as a feature selection technique. It measures the relationship between feature and target variable. If the value if higher, then the feature is of higher importance. ${\chi}^2$ is one of the most efficient feature selection techniques in sentiment analysis. It is calculated as:
\[\tilde{\chi}^2(f) =  \sum_{k=1}^{C}\frac{n_{k}}{N} \times {\chi}^2(f,c_{k})\]
\subsubsection{GSS Coefficient(GSS)}
GSS coefficient was proposed by Galavotti, Sebastiani and Simi \cite{Galavotti}. It is a simplified ${\chi}^2$ function. It has low values for features that are very rare but have high correlation coefficient. Since rarely occurring features are less effective, it eliminates those features and hence frequent features have higher GSS scores. It is measured using the formula:
\begin{equation}
GSS(f) =  \max_{1 \leq k \leq |C|} GSS(f,c_{k}) 
\end{equation}
\begin{equation}
GSS(f,c_{k}) = (n_{f,c_{k}} \times n_{\overline{f},c_{k}})- (n_{f,c_{\overline{k}}}\times n_{\overline{f},c_{k}})
\end{equation}
\subsubsection{Bi-Normal Separation (BNS)}
Bi-Normal Separation of a feature is calculated by measuring the occurrence of the feature in each document. It is the event of random normal variable which exceeds certain threshold. The area under the curve beyond that threshold denotes the prevalence rate. If the feature is more frequent, it means that it's area is larger. It is calculated as:
\begin{equation}
BNS(f) =  \sum_{k=1}^{C}\frac{n_{k}}{N} \times BNS(f,c_{k})
\end{equation}	
\begin{equation}
BNS(f,c_{k}) = (F^{-1} (\frac{n_{f,c_{k}}}{n_{c_{k}}}) - F^{-1}(\frac{n_{f,c_{\overline{k}}}}{n_{c_{\overline{k}}}}))
\end{equation}
\subsubsection{Count Difference (CD)}
Count difference was proposed by Jihong Cai and Fei Song\cite{Cai:Song}. Count difference measures the difference of frequency count of a feature in positive class and negative class. If the CD value is higher then the feature occurs more in positive class. If the value is more negative it means it occurred more in negative class. It is normalized by taking the ratio of count difference to total frequency of the feature. For a binary classification it is calculated as :
\begin{equation}
CD(f) = \left | \frac{n_{f,c_{1}}-n_{f,c_{0}}}{n_{f}} \right |
\end{equation}
\subsubsection{Improved Chi-Square (Imp Chi-Sq)}
Improved Chi-Square was proposed based on Chi-Square for Arabic text classification \cite{Bahassine}. Generally, Chi-Square value is calculated for all the features. Features within the threshold limit are selected irrespective of class label. However, in Improved Chi-Square a constraint is imposed on the number of features to be selected for each class. Hence, features selected will be fairly represented across all the classes.
\subsubsection{Multivariate Relative Discriminative Criterion (MRDC)}
Multivariate Relative Discriminative Criterion (MRDC) for a feature is measured based on the Relative discriminative criterion (RDC) \cite{Rehman} and correlation between features \cite{Labani}. 
Correlation between feature $w_{i}$ and $w_{j}$ is:
\begin{equation}
    correlation(w_{i},w_{j}) = \left | \frac{\sum_{d=1}^{|n|} (f_{i,d}-\overline{f_{i}})(f_{j,d}-\overline{f_{j}})}
    {\sqrt{\sum_{d=1}^{|n|} (f_{i,d}-\overline{f_{i}})^{2}} \sqrt{\sum_{d=1}^{|n|} (f_{j,d}-\overline{f_{j}})^{2}}}\right |
\end{equation}
$f_{i}$, $f_{j}$ are the count vectors of features $w_{i}$ and $w_{j}$ respectively for all the sentences. $\overline{f_{i}}$ and $\overline{f_{j}}$ are the mean values of feature vectors. MRDC for a feature $w_{i}$ is measured as:
\begin{equation}
    MRDC(w_{i}) = RDC(w_{i}) - \sum_{w_{i}\neq w_{j}, w_{j} \in S} correlation(w_{i},w_{j})
\end{equation}
\subsection{Classification}
The features selected after applying these feature selection techniques are trained using machine learning classifiers. Logistic Regression (LR), Support Vector Machines with RBF (SVM-RBF) and Linear Kernels (SVM-L), Decision Trees (DT), Multinomial Naive Bayes (MNB), Bernoulli Naive Bayes (BNB) are the classifiers used for performing sentiment classification.
\subsubsection{Logistic Regression(LR)}
Logistic Regression is a supervised classifier and it is a variant of linear regression. It calculates the linear product of the weight parameters $(W)$ and the input variables $(X)$. It then applies sigmoid function on the linear product and calculates the probability. If the probability is greater than 0.5, it assigns label \say{1} else it assigns label as \say{0}. It works by finding the best set of weight parameters minimizing the loss function. LR is the building block of Multilayer perceptron. 
\begin{eqnarray}
P(y=1|x) = h_{\theta}(x) = \frac{1}{1+e^{-{\theta}^{T} x}}
\end{eqnarray} 
Cost Function to train LR is:
\begin{equation}
    J(\theta) = \frac{-1}{m} \sum y \log h_{\theta}(x) + (1-y) \log (1-h_{\theta}(x))
\end{equation}
\subsubsection{Support Vector Machines}
Support Vector Machines are one of the most powerful supervised machine learning classification algorithms frequently used in image classification and text classification. It transforms the data points into a higher dimensional space so that the data points can be separated linearly. Its builds model by finding a hyperplane that best separates the points. Its main objective is to reduce the sum of the distances of all the data points to that hyperplane. Functions called kernels, decides the hyperplanes to be chosen. If the data can be linearly separated by hyperplane then linear kernel is used. For non-linear data, Radial Basis Function (RBF) kernel is used. The cost function to train a SVM is:
\begin{equation}
    J(w,b) = min_{w,b} \frac{1}{2} ||w||^{2} 
\end{equation}
\begin{equation}
     y^{(i)} (w^{T} x^{(i)} + b) \geq 1, i = 1,2,...,m
\end{equation}
\subsubsection{Decision Tree}
Decision Tree works by constructing tree with internal nodes containing conditions and the leaf nodes containing the class labels. It selects the features to split by calculating information gain and entropy on the feature. The feature with maximum information gain is selected because it provides maximum amount of information when the split is done on the feature. When new data is to be labelled, it is passed to all the internal nodes finally leading to the leaf node. The majority label in that leaf node is assigned to that data.
\subsubsection{Naive Bayes}
Naive Bayes is one of the most frequently used classification algorithms for text classification. Naive Bayes algorithm is based on the bayes probability theorem. There are different variants of naive bayes like Gaussian Naive Bayes (GNB), Multinomial Naive Bayes (MNB) and Bernoulli Naive Bayes (BNB) based on the distributions. Let $f^{(i)} \in \mathbb{R}^{V}$ be the feature count vector for $i^{th}$ training sample where $V$ is the feature set. Let $f_{j}^{(i)}$ be the count of $V_{j}$ for training case $i$. Generally in MNB, the feature vector for input sentence $x_{(k)}$ is $f^{(k)}$ whereas for BNB $x^{(k)}$ =  $\overline{f^{(k)}}$ = 1 ($f^{(k)} > 1$). In BNB, for each feature word in the sentence, if the frequency of the feature word is greater than 1, feature value is taken as 1 and for frequency count 0, it is taken 0.  Hence, BNB is also called Boolean Naive Bayes. However for MNB, we find that binarizing features showed better results \cite{Metsis}. Hence for MNB, if feature is present in a sentence, the feature value is 1 else it is 0.
\subsection{Ensemble Techniques}
\subsubsection{Bagging Ensemble (Bag)}
It is a bootstrap ensemble that creates subsets of data from the original data with replacement. Base Classifiers are trained on each of the data subsets and the individual predictions are combined to output the final prediction. Bagging helps in improving the performance of a weak classifier by training multiple weak classifiers on subsets of original data. 
\subsubsection{Random Subspace Ensemble (RS)}
Let $F$ = $\{f_{1},f_{2},f_{3},.....,f_{n}\}$ denote the feature set. For training $C$ classifiers, $C$ subsets of features are collected with size of each subset being $M$. Base Classifiers are trained on all the subsets and the predictions are combined using voting method or average to output the final prediction. When the data is small and the number of features are large, RS avoids overfitting by selecting subsets of features thereby reducing feature dimensionality.
\section{Experiments}
In this section, we analyze the performance of different FS techniques for sentiment analysis. For the experiments, we used Amazon Reviews Dataset (ARD), IMDB Reviews Dataset (IRD), Yelp Reviews Datasets (YRD) \cite{Kotzias}. The statistics of these datasets are shown in table \ref{tab:1}. Features selected using FS techniques are trained using Logistic Regression (LR), Support Vector Machines with RBF and Linear Kernels, Decision Tree, Multinomial Naive Bayes (MNB) and Bernoulli Naive Bayes (BNB). These classifiers are further trained using Ensemble methods Bagging and Random Subspace.

To evaluate the performance of the FS techniques, Accuracy and F1-score are used as the metrics. We used both metrics to measure the effectiveness of the feature selection techniques. 
Accuracy is the ratio of number of correctly classified samples to the total number of samples. It is defined as:
\begin{eqnarray}
Accuracy = \frac{TP+TN}{(TP+FP+FN+TN)}
\end{eqnarray}
where TP denotes number of True positives, FP denotes False Positives, FN denotes False negatives, TN denotes True Negatives.
Accuracy alone sometimes might be ineffective in classification problems because accuracy doesn't take into account of False positives. Therefore F1-score is used. It is the harmonic mean of precision and recall.
F1 score is defined as:
\begin{eqnarray}
F1-Score = \frac{2*precision*recall}{(precision+recall)}
\end{eqnarray}
Where Precision is the ratio of number of True Positives to number of True Positives and False Positives. Recall is the ratio of number of True Positives to number of True Positives and the number of False Positives.
\begin{eqnarray}
precision = \frac{True positives}{True positives+False positives}
\end{eqnarray}
\begin{eqnarray}
recall = \frac{True positives}{True positives+False negatives}
\end{eqnarray}
\begin{table}
	\centering
	\caption{Statistics of the Datasets}
	\label{tab:1}
		\begin{tabular}{c c c c c c}
		\noalign{\smallskip}\hline\noalign{\smallskip}
		Dataset & Positive & Negative & Train & Test \\
		\hline\noalign{\smallskip}
		Amazon & 500 & 500 & 700 & 300 \\
		\hline\noalign{\smallskip}
		IMDB & 500 & 500 & 700 & 300 \\
		\hline\noalign{\smallskip}
		Yelp & 500 & 500 & 700 & 300 \\
		\hline\noalign{\smallskip}
		MR & 5331 & 5331 & 7108 & 3554 \\
		\hline\noalign{\smallskip}
		Books & 1000 & 1000 & 1600 & 400 \\
		\hline\noalign{\smallskip}
		DVD & 1000 & 1000 & 1600 & 400 \\
		\hline\noalign{\smallskip}
		Electronics & 1000 & 1000 & 1600 & 400 \\
		\hline\noalign{\smallskip}
		Kitchen & 1000 & 1000 & 1600 & 400 \\
		\hline\noalign{\smallskip}
		\end{tabular}
\end{table}
\section{Results and Discussion}

\begin{table}
	\centering
	\caption{Accuracy Scores of FS techniques trained using Base Classifiers for ARD}
	\label{tab:2}       
	\begin{tabular}{c c c c c c c}
	\noalign{\smallskip}\hline\noalign{\smallskip}
	& \multicolumn{6}{c}{Base Classifiers}  \\
	\noalign{\smallskip}\hline\noalign{\smallskip}
		FS Techniques & LR & SVM-RBF & SVM-Linear&  DT & MNB & BNB \\
		\hline\noalign{\smallskip}
		OR &71.67  &73  &71.67   &65.67 &70 &70   \\
		\hline\noalign{\smallskip}
		Chi-Sq & 79.67& 78.67&82 &76.33 &\textbf{82.33} &82\\
		\hline\noalign{\smallskip}
		GSS &81 &79.67 &80 &77 &78.67 &79.33 \\
		\hline\noalign{\smallskip}
		BNS &81.33 &76.33 &81.33 &80.67 &80.33 &76.33   \\
		\hline\noalign{\smallskip}
		CD & 80.33& 80.67&79.33 &79.33 &79 &79.33 \\
		\hline\noalign{\smallskip}
		Imp Chi-Sq &78 &79 &78 &74.67 &74.33  &75 \\
		\hline\noalign{\smallskip}
		MRDC &67.33 &67.33 &66.33 &62.67 &68 &67.67 \\
		\hline\noalign{\smallskip}
	\end{tabular}
\end{table}
\begin{table}
	\centering
	\caption{Accuracy Scores of FS techniques trained using Bagging Ensemble Classifiers on ARD}
	\label{tab:3}       
	\begin{tabular}{c c c c c c c}
	\noalign{\smallskip}\hline\noalign{\smallskip}
	& \multicolumn{6}{c}{Bagging Ensemble Classifiers}  \\
	\noalign{\smallskip}\hline\noalign{\smallskip}
		FS Techniques & LR & SVM-RBF & SVM-Linear&  DT & MNB & BNB \\
		
		\hline\noalign{\smallskip}
		OR & 74.33 &73.67  &74.33   &68.67 &71  &73.67 \\
		\hline\noalign{\smallskip}
		Chi-Sq &80.67 &78.67 &82.67 &79 &82.33 &82.33\\
		\hline\noalign{\smallskip}
		GSS &83 &81.33 &83 &80 &80.33 &81.33 \\
		\hline\noalign{\smallskip}
		BNS & 82.33& 79&82.67 &81.67 &81.67 &82   \\
		\hline\noalign{\smallskip}
		CD & \textbf{83.33}&82 &\textbf{83.33} &81.33 &81 &82 \\
		\hline\noalign{\smallskip}
		Imp Chi-Sq &80.67 &80.33 &81.33 &74.33 &75.33  &78.67 \\
		\hline\noalign{\smallskip}
		MRDC & 69.33& 69.67&69.67 &65.67 &69.67 &70.67 \\
		\hline\noalign{\smallskip}
	\end{tabular}
\end{table}

\begin{table}
	\centering
	\caption{Accuracy Scores of FS techniques trained using Random Subspace Ensemble Classifiers on ARD}
	\label{tab:4}       
	\begin{tabular}{c c c c c c c}
	\noalign{\smallskip}\hline\noalign{\smallskip}
	& \multicolumn{6}{c}{Random Subspace Ensemble Classifiers}  \\
	\noalign{\smallskip}\hline\noalign{\smallskip}
		FS Techniques & LR & SVM-RBF & SVM-Linear&  DT & MNB & BNB \\
		\hline\noalign{\smallskip}
		OR &73.67  &74  &74   &68.33 &73.33&74.67   \\
		\hline\noalign{\smallskip}
		Chi-Sq &81 &79.33 & 82.33& 76&83.33 &82\\
		\hline\noalign{\smallskip}
		GSS & 83.33&83 &\textbf{84} &78 &80.67 &81.33 \\
		\hline\noalign{\smallskip}
		BNS &82.33 &81 &82.67 &81.67 &82 &81.67   \\
		\hline\noalign{\smallskip}
		CD &82.67 &81.33 &81.67 &80.67 &80.33 &81.67 \\
		\hline\noalign{\smallskip}
		Imp Chi-Sq & 80.67& 82.67& 82.33&74.66 &76.67  &78.67 \\
		\hline\noalign{\smallskip}
		MRDC & 68.67 & 68.33 & 69.33 & 63.67 & 69.33 & 69 \\
		\hline\noalign{\smallskip}
	\end{tabular}
\end{table}

\begin{table}
	\centering
	\caption{Accuracy Scores of FS techniques trained using Base Classifiers on IRD}
	\label{tab:5}       
	\begin{tabular}{c c c c c c c}
	\noalign{\smallskip}\hline\noalign{\smallskip}
	& \multicolumn{6}{c}{Base Classifiers}  \\
	\noalign{\smallskip}\hline\noalign{\smallskip}
		FS Techniques & LR & SVM-RBF & SVM-Linear&  DT & MNB & BNB \\
		\hline\noalign{\smallskip}
		OR & 62.67 &58  &63.33   &58 & 63.67&63  \\
		\hline\noalign{\smallskip}
		Chi-Sq &76.67 &77 &76.67 &68 &\textbf{78.67} &72\\
		\hline\noalign{\smallskip}
		GSS & 76& 70.67& 70.33&66 &76.33 &71.67 \\
		\hline\noalign{\smallskip}
		BNS & 74& 71.67&71.33 &65.33 &76.33 &74.33   \\
		\hline\noalign{\smallskip}
		CD &75 &70 &70 &69 &77.67 &71.67 \\
		\hline\noalign{\smallskip}
		Imp Chi-Sq & 71.33&68.33 &69.33 &68 &72.67  &69.67 \\
		\hline\noalign{\smallskip}
		MRDC &64.33 &60.33 &61.67 &59 &68.67 &61.33 \\
		\hline\noalign{\smallskip}
	\end{tabular}
\end{table}

\begin{table}
	\centering
	\caption{Accuracy Scores of FS techniques trained using Bagging Ensemble Classifiers on IRD}
	\label{tab:6}       
	\begin{tabular}{c c c c c c c}
	\noalign{\smallskip}\hline\noalign{\smallskip}
	& \multicolumn{6}{c}{Bagging Ensemble Classifiers}  \\
	\noalign{\smallskip}\hline\noalign{\smallskip}
		FS Techniques & LR & SVM-RBF & SVM-Linear&  DT & MNB & BNB \\
		\hline\noalign{\smallskip}
		OR & 66 & 66 & 64.67  &63.33 &67  &66 \\
		\hline\noalign{\smallskip}
		Chi-Sq & 77.67&77.67 &77.33 &70 &\textbf{79.67} &75.33\\
		\hline\noalign{\smallskip}
		GSS & 76.33&73.67 &73 &69 &78.33 &73.33 \\
		\hline\noalign{\smallskip}
		BNS &75.33 &72.67 &75 &66.67 &76.67 &75.33   \\
		\hline\noalign{\smallskip}
		CD &76 &73.67 &75.33 &71.67 &79 &74.33 \\
		\hline\noalign{\smallskip}
		Imp Chi-Sq &78 &69 &70.67 &66.33 &74  &72.33 \\
		\hline\noalign{\smallskip}
		MRDC & 65& 59& 63.67& 61&68.67 &65.33 \\
		\hline\noalign{\smallskip}
	\end{tabular}
\end{table}

\begin{table}
	\centering
	\caption{Accuracy Scores of FS techniques trained using Random Subspace Ensemble Classifiers on IRD}
	\label{tab:7}       
	\begin{tabular}{c c c c c c c}
	\noalign{\smallskip}\hline\noalign{\smallskip}
	& \multicolumn{6}{c}{Random Subspace Ensemble Classifiers}  \\
	\noalign{\smallskip}\hline\noalign{\smallskip}
		FS Techniques & LR & SVM-RBF & SVM-Linear&  DT & MNB & BNB \\
		\hline\noalign{\smallskip}
		OR & 63.67 &65  &63.33 &63.67  &65.33 &64.33   \\
		\hline\noalign{\smallskip}
		Chi-Sq & 77.33&79 &77.67 &72.67 &\textbf{79.67} &75.67\\
		\hline\noalign{\smallskip}
		GSS & 77& 73.67 &74.33 &71.33 &77.33&78.33 \\
		\hline\noalign{\smallskip}
		BNS & 76& 77& 74.67&70.33 &76.67 &74.67   \\
		\hline\noalign{\smallskip}
		CD &76.67 &72.67 &74.33 &70.67 &78.67 &73.67 \\
		\hline\noalign{\smallskip}
		Imp Chi-Sq &72 &71 &71 &68.33 &73.67  &72.33 \\
		\hline\noalign{\smallskip}
		MRDC &65.67 &60.33 &64.67 &60.33 &69.33 & 64\\
		\hline\noalign{\smallskip}
	\end{tabular}
\end{table}

\begin{table}
	\centering
	\caption{Accuracy Scores of FS techniques trained using Base Classifiers on YRD}
	\label{tab:8}       
	\begin{tabular}{c c c c c c c}
	\noalign{\smallskip}\hline\noalign{\smallskip}
	& \multicolumn{6}{c}{Base Classifiers}  \\
	\noalign{\smallskip}\hline\noalign{\smallskip}
		FS Techniques & LR & SVM-RBF & SVM-Linear&  DT & MNB & BNB \\
		\hline\noalign{\smallskip}
		OR & 63.67 &64.33  &63.33 &61.67 &59.33  &64 \\
		\hline\noalign{\smallskip}
		Chi-Sq &79 &77 &79.33 & 71.67& 76&\textbf{80}\\
		\hline\noalign{\smallskip}
		GSS &76.33 &77 &76 &73.33 &77 & 76.33\\
		\hline\noalign{\smallskip}
		BNS &77.67 &76.67 &78.67 &74.33 &75 &75   \\
		\hline\noalign{\smallskip}
		CD &77 &76.67 &75.67 &71 &77.33 &76.67 \\
		\hline\noalign{\smallskip}
		Imp Chi-Sq &76.67 &75.67 &76.33 &69.67 &75.67  &78.67 \\
		\hline\noalign{\smallskip}
		MRDC & 65.33&63 &65 &60 &64.67 &64.67 \\
		\hline\noalign{\smallskip}
	\end{tabular}
\end{table}

\begin{table}
	\centering
	\caption{Accuracy Scores of FS techniques trained using Bagging Ensemble Classifiers on YRD}
	\label{tab:9}       
	\begin{tabular}{c c c c c c c}
	\noalign{\smallskip}\hline\noalign{\smallskip}
	& \multicolumn{6}{c}{Bagging Ensemble Classifiers}  \\
	\noalign{\smallskip}\hline\noalign{\smallskip}
		FS Techniques & LR & SVM-RBF & SVM-Linear&  DT & MNB & BNB \\
		\hline\noalign{\smallskip}
		OR & 67.33 &66.67  &67.67   &62 & 65.33 &66.67 \\
		\hline\noalign{\smallskip}
		Chi-Sq &79 &78.67 &80.67 &75.67 &80.33 &80.33\\
		\hline\noalign{\smallskip}
		GSS & 78.33&77.67 &77.67 &75 &78.67 &\textbf{81} \\
		\hline\noalign{\smallskip}
		BNS &79 &77 &78.67 &75.33 &\textbf{81} &80.67   \\
		\hline\noalign{\smallskip}
		CD & 79& 77&77.67 &73.67 &79.33 &\textbf{81} \\
		\hline\noalign{\smallskip}
		Imp Chi-Sq &78 &77.67 &77.67 &71.10 &77.67  &80.67 \\
		\hline\noalign{\smallskip}
		MRDC &68.33 & 67.33& 67.67&63 &66.67 &66.67 \\
		\hline\noalign{\smallskip}
	\end{tabular}
\end{table}

\begin{table}
	\centering
	\caption{Accuracy Scores of FS techniques trained using Random Subspace Ensemble Classifiers on YRD}
	\label{tab:10}       
	\begin{tabular}{c c c c c c c}
	\noalign{\smallskip}\hline\noalign{\smallskip}
	& \multicolumn{6}{c}{Random Subspace Ensemble Classifiers}  \\
	\noalign{\smallskip}\hline\noalign{\smallskip}
		FS Techniques & LR & SVM-RBF & SVM-Linear&  DT & MNB & BNB \\
		\hline\noalign{\smallskip}
		OR & 68.33 & 67.67 &68.67   &63.33 &66.67  &67 \\
		\hline\noalign{\smallskip}
		Chi-Sq &79.33 &78.67 &80.67 &74.33 &78.67 &80.33\\
		\hline\noalign{\smallskip}
		GSS &80.33 &80.33 &80.33 &73.33 &79 &79 \\
		\hline\noalign{\smallskip}
		BNS &\textbf{81.33} &80.33 &\textbf{81.33} &75 &81 &80.67   \\
		\hline\noalign{\smallskip}
		CD &81 &78.67 &79.33 &73.67 &78.33 &78.33 \\
		\hline\noalign{\smallskip}
		Imp Chi-Sq & 77.67&77.67& 78.67& 73& 77.67&79.33  \\
		\hline\noalign{\smallskip}
		MRDC & 68.33& 68.33&70 &63 &67.67 &66.67 \\
		\hline\noalign{\smallskip}
	\end{tabular}
\end{table}

\begin{table}
	\centering
	\caption{F1-Scores of FS techniques trained using Base Classifiers on ARD}
	\label{tab:11}       
	\begin{tabular}{c c c c c c c}
	\noalign{\smallskip}\hline\noalign{\smallskip}
	& \multicolumn{6}{c}{Base Classifiers}  \\
	\noalign{\smallskip}\hline\noalign{\smallskip}
		FS Techniques & LR & SVM-RBF & SVM-Linear&  DT & MNB & BNB \\
		\hline\noalign{\smallskip}
		OR & 69.31 &69.66  &68.63   &65.52 &70.20  &69.39 \\
		\hline\noalign{\smallskip}
		Chi-Sq & 78.89& 76.47&81.38 &75.61 &\textbf{82.74} &82.35\\
		\hline\noalign{\smallskip}
		GSS & 81.06& 78.89&80.26 &78.37 &79.87 &80.38 \\
		\hline\noalign{\smallskip}
		BNS & 80.28&73.41 &80.28 &79.43 &80.27 &79.3   \\
		\hline\noalign{\smallskip}
		CD &80.27 &79.86 &79.61 &79.05 &80.37 &80.38 \\
		\hline\noalign{\smallskip}
		Imp Chi-Sq & 78.15& 77.90&76.92 &75.48 &76.16  &76.20 \\
		\hline\noalign{\smallskip}
		MRDC &63.97 &60.80 &63.00 &57.25 &65.71 &65.23 \\
		\hline\noalign{\smallskip}
	\end{tabular}
\end{table}
\begin{table}
	\centering
	\caption{F1-Scores of FS techniques trained using Bagging Ensemble Classifiers on ARD}
	\label{tab:12}       
	\begin{tabular}{c c c c c c c}
	\noalign{\smallskip}\hline\noalign{\smallskip}
	& \multicolumn{6}{c}{Bagging Ensemble Classifiers}  \\
	\noalign{\smallskip}\hline\noalign{\smallskip}
		FS Techniques & LR & SVM-RBF & SVM-Linear&  DT & MNB & BNB \\
		\hline\noalign{\smallskip}
		OR & 71.59 &70.41 &72  &69.82  &75.21  &72.66 \\
		\hline\noalign{\smallskip}
		Chi-Sq & 80.67 &76.81 &81.43  & 78.2 &82.96  &82.74\\
		\hline\noalign{\smallskip}
		GSS & 82.71 &80.14 &83.28  &79.87  & 82.39 &81.60 \\
		\hline\noalign{\smallskip}
		BNS & 81 &77.26 &82.07  &80.7  &81.73  &81.76   \\
		\hline\noalign{\smallskip}
		CD & 83.11 & 81.38& \textbf{83.44} &80.42  &82.67  &82.80 \\
		\hline\noalign{\smallskip}
		Imp Chi-Sq & 80.14 &79.44 &80.82  &75.241  &77.01  &79.08 \\
		\hline\noalign{\smallskip}
		MRDC & 65.41 &71.11 &66.42  &62.82  &74.51  &74.57 \\
		\hline\noalign{\smallskip}
	\end{tabular}
\end{table}

\begin{table}
	\centering
	\caption{F1-Scores of FS techniques trained using Random Subspace Ensemble Classifiers on ARD}
	\label{tab:13}       
	\begin{tabular}{c c c c c c c}
	\noalign{\smallskip}\hline\noalign{\smallskip}
	& \multicolumn{6}{c}{Random Subspace Ensemble Classifiers}  \\
	\noalign{\smallskip}\hline\noalign{\smallskip}
		FS Techniques & LR & SVM-RBF & SVM-Linear&  DT & MNB & BNB \\
		\hline\noalign{\smallskip}
		OR & 72.28 &70.68 &71.53  &69.45  &72.60  &73.43 \\
		\hline\noalign{\smallskip}
		Chi-Sq & 80.55 &77.86 &81.66  &76.92  & 83.77 &82.58\\
		\hline\noalign{\smallskip}
		GSS & 83.77 &82.22 &\textbf{84}  &79.99  &81.76  &82.28 \\
		\hline\noalign{\smallskip}
		BNS & 81.40 &79.86 &82.07  &81.10  &81.88  &81.48   \\
		\hline\noalign{\smallskip}
		CD & 82.89 &80.08 & 81.61 &80.54  &81.50  &82.32 \\
		\hline\noalign{\smallskip}
		Imp Chi-Sq & 80.67 &81.56 & 81.27 &75.64  &78.53  &79.22 \\
		\hline\noalign{\smallskip}
		MRDC & 64.93 &62.75 & 64.34 &69.47  & 66.91 & 66.67\\
		\hline\noalign{\smallskip}
	\end{tabular}
\end{table}

\begin{table}
	\centering
	\caption{F1-Scores of FS techniques trained using Base Classifiers on IRD}
	\label{tab:14}       
	\begin{tabular}{c c c c c c c}
	\noalign{\smallskip}\hline\noalign{\smallskip}
	& \multicolumn{6}{c}{Base Classifiers}  \\
	\noalign{\smallskip}\hline\noalign{\smallskip}
		FS Techniques & LR & SVM-RBF & SVM-Linear&  DT & MNB & BNB \\
		\hline\noalign{\smallskip}
		OR & 60.28 &57.89 &53.33  &60.71  &62.28  &58.11 \\
		\hline\noalign{\smallskip}
		Chi-Sq & 76.97 &77.67 &77.99  & 72.41 &\textbf{79.08}  &66.93\\
		\hline\noalign{\smallskip}
		GSS & 75.34 & 67.16&70.43  &71.5  &77.17  &66.4 \\
		\hline\noalign{\smallskip}
		BNS & 72.54 &69.09 &70.34  & 60.61 &76.25  &72.59   \\
		\hline\noalign{\smallskip}
		CD & 74.40 &66.17 &70.19  &74.38  &78.46  &66.40 \\
		\hline\noalign{\smallskip}
		Imp Chi-Sq & 70.75 &66.43 &68.92  &64.71  &73.55  &67.15 \\
		\hline\noalign{\smallskip}
		MRDC & 59.93 &53.70 & 55.25 & 49.8 &67.36  &50.85 \\
		\hline\noalign{\smallskip}
	\end{tabular}
\end{table}

\begin{table}
	\centering
	\caption{F1-Scores of FS techniques trained using Bagging Ensemble Classifiers on IRD}
	\label{tab:15}       
	\begin{tabular}{c c c c c c c}
	\noalign{\smallskip}\hline\noalign{\smallskip}
	& \multicolumn{6}{c}{Bagging Ensemble Classifiers}  \\
	\noalign{\smallskip}\hline\noalign{\smallskip}
		FS Techniques & LR & SVM-RBF & SVM-Linear&  DT & MNB & BNB \\
		\hline\noalign{\smallskip}
		OR & 63.31 &65.07 & 61.87 & 65.19 &68.97  &61.36 \\
		\hline\noalign{\smallskip}
		Chi-Sq & 78.18 & 78.18&78.21  &73.37  &79.87  &72.59\\
		\hline\noalign{\smallskip}
		GSS & 76.09 &73.22 &72.35  &72.57  &78.69  &68.75 \\
		\hline\noalign{\smallskip}
		BNS & 73.19 & 70.71& 73.31 &63.50  &76.35  &74.13   \\
		\hline\noalign{\smallskip}
		CD & 76.32 &72.47 &73.19  &75.22  &\textbf{80.25}  &70.27 \\
		\hline\noalign{\smallskip}
		Imp Chi-Sq & 71.82 &68.04 &69.44  &63.8  &74.51  & 69.82\\
		\hline\noalign{\smallskip}
		MRDC &63.16  &52.87 & 56.57 &54.83  &71.17  &59.69 \\
		\hline\noalign{\smallskip}
	\end{tabular}
\end{table}

\begin{table}
	\centering
	\caption{F1-Scores of FS techniques trained using Random Subspace Ensemble Classifiers on IRD}
	\label{tab:16}       
	\begin{tabular}{c c c c c c c}
	\noalign{\smallskip}\hline\noalign{\smallskip}
	& \multicolumn{6}{c}{Random Subspace Ensemble Classifiers}  \\
	\noalign{\smallskip}\hline\noalign{\smallskip}
		FS Techniques & LR & SVM-RBF & SVM-Linear&  DT & MNB & BNB \\
		\hline\noalign{\smallskip}
		OR & 61.48 & 59.77&58.96  &61.75  &63.89  &61.65 \\
		\hline\noalign{\smallskip}
		Chi-Sq & 77.92 &79.61 &78.59  & 75.45 & \textbf{80.39} &73.45\\
		\hline\noalign{\smallskip}
		GSS & 76.92 & 71.27&74.42  &74.85  &78.06  &77.19 \\
		\hline\noalign{\smallskip}
		BNS & 75 &68.82 &72.46  &67.39  &76.67  &73.05   \\
		\hline\noalign{\smallskip}
		CD & 75.52 &70.07 &72.59  &74.7  &79.49  &69.73 \\
		\hline\noalign{\smallskip}
		Imp Chi-Sq & 72 &69.04 &70.10  &  67.13&74.1  &68.68 \\
		\hline\noalign{\smallskip}
		MRDC & 61.99 &54.41 &57.49  &54.05  &68.28  &57.14 \\
		\hline\noalign{\smallskip}
	\end{tabular}
\end{table}
\begin{table}
	\centering
	\caption{F1-Scores of FS techniques trained using Base Classifiers on YRD}
	\label{tab:17}       
	\begin{tabular}{c c c c c c c}
	\noalign{\smallskip}\hline\noalign{\smallskip}
	& \multicolumn{6}{c}{Base Classifiers}  \\
	\noalign{\smallskip}\hline\noalign{\smallskip}
		FS Techniques & LR & SVM-RBF & SVM-Linear&  DT & MNB & BNB \\
		\hline\noalign{\smallskip}
		OR &59.17  & 54.08& 54.54 &59.36  &62.11  &58.46 \\
		\hline\noalign{\smallskip}
		Chi-Sq & 77.89 &74.54 &78.10  &69.53  &77.78  & \textbf{79.59}\\
		\hline\noalign{\smallskip}
		GSS & 73.99 &74.35 &73.91  &72.03  &78.09  &75.93 \\
		\hline\noalign{\smallskip}
		BNS & 75.09 &72.44 & 76.98 & 72 &77.20  &77.20   \\
		\hline\noalign{\smallskip}
		CD & 74.73 &72.87 &73.45  &70.10  &\textbf{78.84}  &76.35 \\
		\hline\noalign{\smallskip}
		Imp Chi-Sq & 74.26 &71.81 &73.99  &66.67  &76.83  &77.30 \\
		\hline\noalign{\smallskip}
		MRDC & 60.61 & 53.94& 58.80 &53.85  &68.45  &68.45 \\
		\hline\noalign{\smallskip}
	\end{tabular}
\end{table}

\begin{table}
	\centering
	\caption{F1-Scores of FS techniques trained using Bagging Ensemble Classifiers on YRD}
	\label{tab:18}       
	\begin{tabular}{c c c c c c c}
	\noalign{\smallskip}\hline\noalign{\smallskip}
	& \multicolumn{6}{c}{Bagging Ensemble Classifiers}  \\
	\noalign{\smallskip}\hline\noalign{\smallskip}
		FS Techniques & LR & SVM-RBF & SVM-Linear&  DT & MNB & BNB \\
		\hline\noalign{\smallskip}
		OR & 62.88 &61.24 &60.08  &60.42  &68.09  &62.41 \\
		\hline\noalign{\smallskip}
		Chi-Sq & 78.64 &76.81 & 79.58 &74.74  &79.99  &80.13\\
		\hline\noalign{\smallskip}
		GSS & 76.36 &75.81 &75.63  &73.68  &80  &80 \\
		\hline\noalign{\smallskip}
		BNS & 76.22 &73.15 & 76.29 &72.18  & 80 &79.27   \\
		\hline\noalign{\smallskip}
		CD &77.09  &72.94 &75.09  &72.28  & \textbf{80.63} &80.14 \\
		\hline\noalign{\smallskip}
		Imp Chi-Sq & 76.26 & 74.52& 75.64 &71.08  &78.86  &79.72 \\
		\hline\noalign{\smallskip}
		MRDC & 65.20 &61.41 & 62.55 &57.99  &70.06  &70.41 \\
		\hline\noalign{\smallskip}
	\end{tabular}
\end{table}

\begin{table}
	\centering
	\caption{F1-Scores of FS techniques trained using Random Subspace Ensemble Classifiers on YRD}
	\label{tab:19}       
	\begin{tabular}{c c c c c c c}
	\noalign{\smallskip}\hline\noalign{\smallskip}
	& \multicolumn{6}{c}{Random Subspace Ensemble Classifiers}  \\
	\noalign{\smallskip}\hline\noalign{\smallskip}
		FS Techniques & LR & SVM-RBF & SVM-Linear&  DT & MNB & BNB \\
		\hline\noalign{\smallskip}
		OR & 62.75 &59.76 &62.99  &61.54  &65.28  &62.92 \\
		\hline\noalign{\smallskip}
		Chi-Sq & 78.62 &77.30 &80  &73.72  &78.23  &80.27\\
		\hline\noalign{\smallskip}
		GSS & 80.13 &79.44 &79.15  &73.15  & 80.73 &78.93 \\
		\hline\noalign{\smallskip}
		BNS & 80 &78.39 &79.56  &72.92  &80.13  & 79.58  \\
		\hline\noalign{\smallskip}
		CD & 79.86 &76.64 &\textbf{80.75}  &73.93  &80  & 79.62\\
		\hline\noalign{\smallskip}
		Imp Chi-Sq & 77.29 &74.91 &75.94  &71.17  &79.13  & 78.47\\
		\hline\noalign{\smallskip}
		MRDC & 63.88 &64.15 &65.91  &57.47  &71.04  &70.24 \\
		\hline\noalign{\smallskip}
	\end{tabular}
\end{table}
Tables \ref{tab:2}, \ref{tab:3}, \ref{tab:4} show the accuracy scores of FS techniques trained using Base Classifiers, Bagging and Random Subspace on Amazon dataset respectively. Tables \ref{tab:5}, \ref{tab:6}, \ref{tab:7} show the accuracy scores of FS techniques trained using Base Classifiers, Bagging and Random Subspace on IMDB dataset respectively. Tables \ref{tab:8}, \ref{tab:9}, \ref{tab:10} show the accuracy scores of FS techniques trained using Base Classifiers, Bagging and Random Subspace on Yelp dataset respectively.
Tables \ref{tab:11}, \ref{tab:12}, \ref{tab:13} show the F1-scores of FS techniques trained using Base Classifiers, Bagging and Random Subspace on Amazon dataset respectively. Tables \ref{tab:14}, \ref{tab:15}, \ref{tab:16} show the F1-scores of FS techniques trained using Base Classifiers, Bagging and Random Subspace on IMDB dataset respectively. Tables \ref{tab:17}, \ref{tab:18}, \ref{tab:19} show the F1-scores of FS techniques trained using Base Classifiers, Bagging and Random Subspace on Yelp dataset respectively.  
Results show that Ensemble classifiers achieved significantly better results than base classifiers.
\subsection{Comparison among FS techniques:}
Among the FS techniques, Chi-Square (Chi) and Count Difference (CD) performed better than other FS techniques. Count Difference takes the relative frequencies between classes. Therefore, if a feature occurs frequently in one class compared to other, it's CD value will be high. It means that the feature is important in sentiment analysis. Also, if a feature is present in both the classes with less frequency difference, it doesn't help much in distinguishing between classes. The CD value will be less for such features, thereby neglecting the feature which is of limited importance. Chi-Square test helps in finding the dependency of a feature on a the class. If the Chi-Square value is less, feature is much likely to be independent of class. Such feature, contribute very little in classification.  
\subsection{Comparison among Base Classifiers:} \label{sssec:classifers}
Multinomial Naive Bayes and Logistic Regression achieved significantly better performance than other base classifiers.  
\subsection{Comparison among Ensemble methods:}
Random Subspace performed better than Bagging in sentiment classification. In Bagging, classifiers are trained on subsets of data but with the same feature set as original data. This can sometimes lead to overfitting. In case of Random Subspace, Classifiers are trained on feature subsets and hence classifiers performance will be better.
\section{Comparison with Neural Networks}
We compare the best performing Feature Selection techniques as discussed in section \ref{sssec:classifers} with some of the neural network architectures\footnotemark\footnotetext{Code will be available at repository https://github.com/avinashsai/MTAP}.

\subsection{Datasets}
We used the following datasets for comparison between Neural Networks and FS techniques. 
\subsubsection{Movie Reviews (MR)}
Movie Reviews (MR) is a binary sentiment classification dataset \cite{Pang:Lee}. It has 5331 positive and 5331 negative reviews. We used the training and test data split as used by Liu \cite{Liu:Qiu}.
\subsection{Multi Domain Dataset}
Multi Domain Dataset consists of review sentences from 4 domains Books, DVD, Electronics, Kitchen. Each domain has 2000 reviews with 1000 positive and 1000 negative polarity reviews\footnotemark\footnotetext{Train,Test splits can found in https://github.com/avinashsai/Cross-domain-sentiment-analysis/tree/master/Dataset/Actualdata}. 
\subsection{Comparison Methods}
\subsubsection{FastText+LSTM}
FastText \cite{Joulin} is a distributional word vector model, which treats the word embedding as the average of word and its n-grams. FastText model is pretrained on training data and then fed into a LSTM network.
\subsubsection{Paragraph2Vec}
Paragraph2Vec (PV) \cite{Le:Mikolov} is similar to Word2Vec model except that Paragraph2Vec outputs a single vector for each paragraph. We train Paragraph2Vec on training data and Feed Forward Neural Network is used as the classifier. We considered DBOW and DM models for training Paragraph2Vec.
\subsubsection{LSTM}
LSTM is a strong baseline for sentiment analysis. We used GloVe pretrained word vectors\footnotemark\footnotetext{http;//nlp.stanford.edu/data/glove.840B.300d.zip} as the input to LSTM.
\subsubsection{Bi-LSTM}
In Bi-LSTM, one LSTM is trained from the beginning of the sentence and another LSTM is trained from the end of the sentence. The final hidden states of both LSTMs' are concatenated for final prediction. 
\subsubsection{Convolutional Neural Network (CNN)}
CNN achieved state of the art results in text classification \cite{Kim}. We examined CNN-rand \cite{Kim} which uses randomly initialized word embeddings as input and CNN-static \cite{Kim} which uses pretrained word vectors as input and are not updated in training.
\subsubsection{Transformable Convolutional Neural Network (TF)}
It is a variant of CNN, proposed, to provide CNNs the flexibility to learn complex features and adaptability for transformation of features in text classification \cite{Xiao}. TF-DCNN and TF-MCCNN are the two models used as baselines. \\

Table \ref{tab:20}, \ref{tab:21} shows the performance comparison of neural network architectures with feature selection techniques. Results show that feature selection methods trained using Ensemble classifiers achieved better results than neural network architectures. 

%%%%%%%%%%%%%%%%%%%%%%%%%%%%%%%%%%%%%%%%%%%%%%%%%%%%%%%%%%%%%%%%%%%
\begin{figure}
\centering
	\includegraphics[width=80mm]{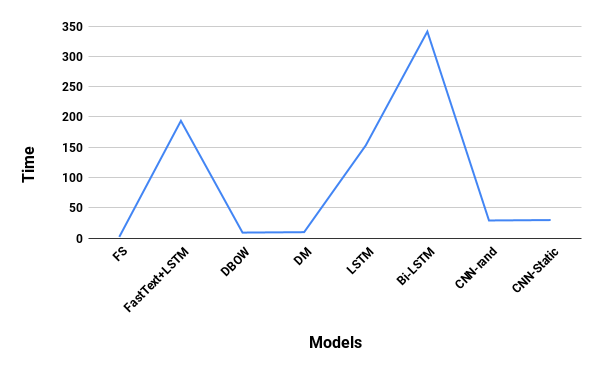}
	\caption{Training times for models on Books dataset}
	\label{fig:2}       
\end{figure}
\begin{figure}
\centering
	\includegraphics[width=70mm]{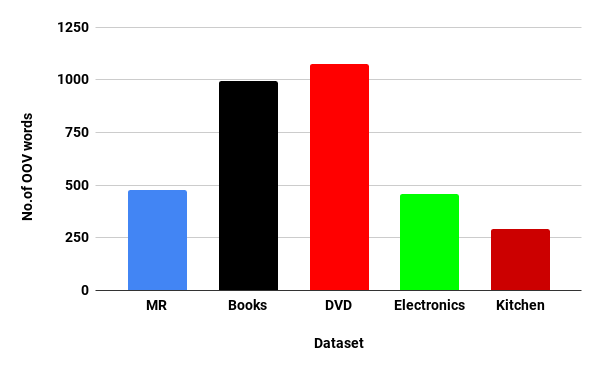}
	\caption{Number of OOV words for each dataset}
	\label{fig:3}       
\end{figure}
\begin{table}
   \setlength{\tabcolsep}{14pt}
	\centering
	\caption{Accuracy scores on sentiment analysis datasets. Every neural network model is ran 5 times and averaged result is reported.}
	\label{tab:20}       
	\begin{tabular}{c c c c c c}
	\hline\noalign{\smallskip}
	Model & MR & Books & DVD & Electronics & Kitchen \\
	\hline\noalign{\smallskip}
	Fasttext &53.65 &49.15 &48.99 &54.39 &52.54 \\
	\hline\noalign{\smallskip}
	PV-DBOW & 58.3& 69.6& 69.25&69.1 &73.44 \\
	\hline\noalign{\smallskip}
	PV-DM &58.08 &70.24 &69.10 &68.3 &72.6\\
	\hline\noalign{\smallskip}
	LSTM & 75.97&72.55 &72.3 &72.55 &78.7\\
	\hline\noalign{\smallskip}
	Bi-LSTM & 75.64& 72.45& 74.8& 72.45&79.35 \\
	\hline\noalign{\smallskip}
	CNN-Rand & 73.20& 74.1& 71.2& 74.1& 81.45\\
	\hline\noalign{\smallskip}
	CNN-Static &73.20 & 74.04&71 &74.04 &81.35\\
	\hline\noalign{\smallskip}
	TF-DCNN \cite{Xiao} & \textbf{$-$} & \textbf{82.0}&81.3 & 81.7 & 81.2\\
	\hline\noalign{\smallskip}
	TF-MCCNN \cite{Xiao} & \textbf{$-$} & 81 &81.6 & 81.1 & 81.4\\
	\hline\noalign{\smallskip}
	CD+LR+Bag & &75.09 &79.75 & \textbf{87.25}&82.75\\
	\hline\noalign{\smallskip}
	CD+LR+RS & 75.63& 75.63&82.25 & \textbf{87.25} &\textbf{84.25}\\
	\hline\noalign{\smallskip}
	CD+MNB+Bag & 75.63& 75.63& 82& 82.75&82\\
	\hline\noalign{\smallskip}
	CD+MNB+RS & 75.54&75.55 &81.5 & 83.25&82.75\\
	\hline\noalign{\smallskip}
	Chi+LR+Bag &75.40 & 79& 83.5&84.75 &84\\
	\hline\noalign{\smallskip}
	Chi+LR+RS & 75.46& 80& 83.25&84.5 &83.5\\
	\hline\noalign{\smallskip}
	Chi+MNB+Bag &76.36 & 79.75&\textbf{84.5} & 84.75 &83.75\\
	\hline\noalign{\smallskip}
	Chi+MNB+RS & \textbf{76.47}& 79.75&84 & 84.75&83.75\\
	\hline\noalign{\smallskip}
	\end{tabular}
\end{table}
%%%%%%%%%%%%%%%%%%%%%%%%%%%%%%%%%%%%%%%%%%%%%%%%%%%%%%%%%%%%%%%%%%%%%%%%%%%%%
\begin{table}
   \setlength{\tabcolsep}{14pt}
	\centering
	\caption{F1-Scores on sentiment analysis datasets. Every neural network model is ran 5 times and averaged result is reported.}
	\label{tab:21}       
	\begin{tabular}{c c c c c c}
	\hline\noalign{\smallskip}
	Model & MR & Books & DVD & Electronics & Kitchen \\
	\hline\noalign{\smallskip}
	Fasttext & 49.42&51.19 &45.44 &55.07 &50.75  \\
	\hline\noalign{\smallskip}
	PV-DBOW & 59.60&70.47 &71.35 &70.63 &73.54 \\
	\hline\noalign{\smallskip}
	PV-DM & 58.85& 70.96&70.54 &69.81 &72.75\\
	\hline\noalign{\smallskip}
	LSTM & 75.90&73.21 &73.65 &73.21 &78.77\\
	\hline\noalign{\smallskip}
	Bi-LSTM &75.89 &73.64 &75.01 &73.64 &79.33 \\
	\hline\noalign{\smallskip}
	CNN-Rand & 73.51&75.32 &71.01 &75.32 &80.86\\
	\hline\noalign{\smallskip}
	CNN-Static &73.54 & 75.31&70.78 &75.31 &80.74\\
	\hline\noalign{\smallskip}
	TF-DCNN \cite{Xiao} & $-$&$-$ &$-$ &$-$ &$-$\\
	\hline\noalign{\smallskip}
	TF-MCCNN \cite{Xiao} & $-$&$-$ &$-$ &$-$ &$-$\\
	\hline\noalign{\smallskip}
	CD+LR+Bag & 74.85& 74.85&80.57 & \textbf{87.46} &82.26\\
	\hline\noalign{\smallskip}
	CD+LR+RS & 75.49&75.49& 82.64 & \textbf{87.46} & \textbf{83.96}\\
	\hline\noalign{\smallskip}
	CD+MNB+Bag & 75.95& 75.96 & 81.63& 82.79&82.26\\
	\hline\noalign{\smallskip}
	CD+MNB+RS & 75.58&75.58 &81.02& 83.37&82.70\\
	\hline\noalign{\smallskip}
	Chi+LR+Bag & 75.46&78.85 &83.66 &84.86 &83.15\\
	\hline\noalign{\smallskip}
	Chi+LR+RS &75.62 &79.80 &83.69 &84.57 &83.07\\
	\hline\noalign{\smallskip}
	Chi+MNB+Bag &76.70 &\textbf{80.56} &\textbf{84.42} & 85.37 &83.95\\
	\hline\noalign{\smallskip}
	Chi+MNB+RS & \textbf{76.89}&80.09 &83.83 &85.15 &83.87\\
	\hline\noalign{\smallskip}
	\end{tabular}
\end{table}
%%%%%%%%%%%%%%%%%%%%%%%%%%%%%%%%%%%%%%%%%%%%%%%%%%%%%%%%%%%%%%%%%%%%%%%%%%%%%
\begin{table}
   \setlength{\tabcolsep}{14pt}
	\centering
	\caption{Number of Model Parameters on Books Dataset}
	\label{tab:22}
	\begin{tabular}{c c}
	\hline\noalign{\smallskip}
	    Model & No.of Parameters \\
	    \hline\noalign{\smallskip}
	     LR &  8001 \\
	     \hline\noalign{\smallskip}
	     LSTM & 5,652,633\\
	     \hline\noalign{\smallskip}
	     Bi-LSTM & 5,884,033 \\
	     \hline\noalign{\smallskip}
	     CNN-rand & 5,821,801 \\
	     \hline\noalign{\smallskip}
	     CNN-static & 5,821,801 \\
	     \hline\noalign{\smallskip}
	\end{tabular}
	\end{table}

\subsection{Advantages of using FS techniques over Neural Networks:}
FS techniques has the following merits compared to Neural Networks:
\subsubsection{Handling  Out-of-Vocabulary Words (OOV):}
The input to these neural network architectures are pretrained word embeddings. These are trained on large corpora of nearly 1.6 billion words. However, one of the major problems with these embedding models is the inability to handle unknown or Out-of-Vocabulary Words (OOV). Figure \ref{fig:3} shows the number of OOV words for each dataset. Also, these embedding models suffer from word sense disambiguation in which the same word can have different contextual meanings. For example \say{bank}. For data scarce languages, it is hard to train word2vec models. On the other hand, FS techniques don't have OOV problem because they operates at word level. Also, FS techniques can be used to perform sentiment analysis even with very limited data and hence don't suffer from  data scarcity.
\subsubsection{Training Time and Model Hyperparameters:}
Figure \ref{fig:2} shows the training time for different models. Table \ref{tab:22} shows the number of parameters required to train a model. The time and model parameters required to train FS techniques with Ensemble Classifiers are significantly very less compared to neural network architectures like LSTM, CNN.
\section{Conclusion}
From the experiments, it is evident that Feature Selection techniques achieve excellent results on sentiment analysis. Among FS techniques, Chi-Square and Count Difference showed superior results. Among Base Classifiers, Multinomial Naive Bayes and Logistic Regression demonstrate remarkable performance. In case of Ensemble methods, Random Subspace achieved superior results compared to Bagging. FS techniques trained using Ensemble techniques outperformed neural network architectures with less number of parameters and training time thereby reducing the need for extensive hyper-parameter tuning.

\section{Conflict of Interest:}
The authors declare that they have no conflict of interest.


\begin{thebibliography}{}
\bibitem{Maaten:Postma}
Van Der Maaten, L., Postma, E. and Van den Herik, J., 2009. Dimensionality reduction: a comparative. J Mach Learn Res, 10(66-71), p.13.
\bibitem{Tang:Alelyani}
Tang, J., Alelyani, S. and Liu, H., 2014. Feature selection for classification: A review. Data classification: Algorithms and applications, p.37.
\bibitem{Das}
Das, S., 2001, June. Filters, wrappers and a boosting-based hybrid for feature selection. In Icml (Vol. 1, pp. 74-81).
\bibitem{Robnik:Koneneko}
Robnik-Šikonja, M. and Kononenko, I., 2003. Theoretical and empirical analysis of ReliefF and RReliefF. Machine learning, 53(1-2), pp.23-69.
\bibitem{Plackett}
Plackett, R.L., 1983. Karl Pearson and the chi-squared test. International Statistical Review/Revue Internationale de Statistique, pp.59-72.
\bibitem{Harris}
Harris, Zellig S. "Distributional structure." Word 10.2-3 (1954): 146-162.
\bibitem{Feldman:Ronen}
Feldman,Ronen: Techniques and applications for sentiment analysis, In: Communications of the ACM, Vol 56, No (4) ,pp 82-89 (2013).% Format for books
\bibitem{Dang:Zhang:Chen}
Y. Dang, Y. Zhang, H. Chen: A lexicon-enhanced method for sentiment classification: An experiment on online product reviews, In: Intelligent Systems, IEEE Vol 25, No (4), pp 46-53 (2010).
\bibitem{Hotelling}
 Hotelling H. Analysis of a complex of statistical variables into principal components. J Educ Psychol,25:417–441 (1933).
\bibitem{Anderson:Cafarella}
M. R. Anderson and M. Cafarella: Input selection for fast feature engineering, In: IEEE,32nd International Conference on Data Engineering (ICDE), pp 577-588, (2016).
\bibitem{Byeon:Rasheed}
 B. Byeon and K. Rasheed: Simultaneously Removing Noise and Selecting Relevant Features for High Dimensional Noisy Data, In: Seventh International Conference on Machine Learning and Applications, pp 147-152 (2008).
 \bibitem{Pang:Lee:Vaithyanathan}
 Pang, Bo and Lee, Lillian and Vaithyanathan, Shivakumar: Thumbs Up?: Sentiment Classification Using Machine Learning Techniques, In: Proceedings of the ACL-02 Conference on Empirical Methods in Natural Language Processing - Vol 10,EMNLP '02,pp 79-86 (2002)
 \bibitem{Wang:Manning}
 Wang, S. and Manning, C.D., 2012, July. Baselines and bigrams: Simple, good sentiment and topic classification. In Proceedings of the 50th annual meeting of the association for computational linguistics: Short papers-volume 2 (pp. 90-94). Association for Computational Linguistics.
 \bibitem{Hochreiter:Bengio}
 Hochreiter, S., Bengio, Y., Frasconi, P. and Schmidhuber, J., 2001. Gradient flow in recurrent nets: the difficulty of learning long-term dependencies.
 \bibitem{Pascanu:Mikolov:Bengio}
 Pascanu, R., Mikolov, T. and Bengio, Y., 2012. Understanding the exploding gradient problem. CoRR, abs/1211.5063, 2.
 \bibitem{Hochreiter:Schmidhuber}
 Hochreiter, S. and Schmidhuber, J., 1997.Long short-term memory. Neural computation, 9(8),pp.1735-1780.
 \bibitem{Kim}
 Kim, Y., 2014.Convolutional neural networks for sentence classification. arXiv preprint arXiv:1408.5882.
 \bibitem{Conneau:Lecun}
 Conneau, A., Schwenk, H., Barrault, L. and Lecun, Y., 2016. Very deep convolutional networks for text classification. arXiv preprint arXiv:1606.01781.
 \bibitem{Breiman}
 Breiman, L., 1996. Bagging predictors. Machine learning, 24(2), pp.123-140.
 \bibitem{Barandiaran}
 Barandiaran, I., 1998. The random subspace method for constructing decision forests. IEEE Trans. Pattern Anal. Mach. Intell, 20(8), pp.1-22.
 \bibitem{Morinaga}
 Morinaga, S., Yamanishi, K., Tateishi, K. and Fukushima, T., 2002, July. Mining product reputations on the web. In Proceedings of the eighth ACM SIGKDD international conference on Knowledge discovery and data mining (pp. 341-349). ACM.
 \bibitem{Jones}
 Jones, K.S., 2004. A statistical interpretation of term specificity and its application in retrieval. Journal of documentation.
 \bibitem{Tim:Irena}
 O’Keefe, T. and Koprinska, I., 2009, December. Feature selection and weighting methods in sentiment analysis. In Proceedings of the 14th Australasian document computing symposium, Sydney (pp. 67-74).
 \bibitem{Tan:Zhang}
 Tan, S. and Zhang, J., 2008. An empirical study of sentiment analysis for chinese documents. Expert Systems with applications, 34(4), pp.2622-2629.
 \bibitem{Abbasi:Chen}
 Abbasi A., Chen H.C., and Salem A. (2008). “Sentiment analysis in multiple languages: Feature selection for opinion classification in web forums”. In ACM Transactions on Information Systems (TOIS), 2008. 26(3).
 \bibitem{Agarwal:Mittal:Acl}
 Agarwal, B. and Mittal, N., 2012, December. Categorical probability proportion difference (CPPD): a feature selection method for sentiment classification. In Proceedings of the 2nd Workshop on Sentiment Analysis where AI meets Psychology (pp. 17-26).
 \bibitem{Wang:Li:Wei}
 Wang S., Li D., Wei Y., Li H. (2009) A Feature Selection Method Based on Fisher’s Discriminant Ratio for Text Sentiment Classification. In: Liu W., Luo X., Wang F.L., Lei J. (eds) Web Information Systems and Mining. WISM 2009. Lecture Notes in Computer Science, vol 5854. Springer, Berlin, Heidelberg.
 \bibitem{Agarwal:Mittal}
 Agarwal, B. and Mittal, N., 2013, March. Optimal feature selection for sentiment analysis. In International Conference on Intelligent Text Processing and Computational Linguistics (pp. 13-24). Springer, Berlin, Heidelberg.
 \bibitem{Pong:Kaewmak}
 C. Pong-Inwong and K. Kaewmak : Improved sentiment analysis for teaching evaluation using feature selection and voting ensemble learning integration, In: 2nd IEEE International Conference on Computer and Communications (ICCC), pp 1222-1225 (2016).
 \bibitem{Chi:Siew}
 X. Chi and T. P. Siew and E. Cambria: Adaptive two-stage feature selection for sentiment classification, In: IEEE International Conference on Systems, Man, and Cybernetics (SMC), pp 1238-1243 (2017).
 \bibitem{Bahassine}
 Bahassine, S., Madani, A., Al-Sarem, M. and Kissi, M., 2018. Feature selection using an improved Chi-square for Arabic text classification. Journal of King Saud University-Computer and Information Sciences.
 \bibitem{Oussous}
 Oussous, A., Lahcen, A.A. and Belfkih, S., 2019, March. Impact of Text Pre-processing and Ensemble Learning on Arabic Sentiment Analysis. In Proceedings of the 2nd International Conference on Networking, Information Systems and Security (p. 65). ACM.
 \bibitem{Lopez}
 López, M., Valdivia, A., Martínez-Cámara, E., Luzón, M.V. and Herrera, F., 2019. E2SAM: Evolutionary ensemble of sentiment analysis methods for domain adaptation. Information Sciences, 480, pp.273-286.
 \bibitem{Lee:Yu}
 Lee, J., Yu, I., Park, J. and Kim, D.W., 2019. Memetic feature selection for multilabel text categorization using label frequency difference. Information Sciences, 485, pp.263-280.
 \bibitem{Labani}
 Labani, M., Moradi, P., Ahmadizar, F. and Jalili, M., 2018. A novel multivariate filter method for feature selection in text classification problems. Engineering Applications of Artificial Intelligence, 70, pp.25-37.
 \bibitem{Abdi}
 Abdi, A., Shamsuddin, S.M., Hasan, S. and Piran, J., 2019. Deep learning-based sentiment classification of evaluative text based on Multi-feature fusion. Information Processing and Management, 56(4), pp.1245-1259.
 \bibitem{Gao:Xuan}
 Gao, Z., Xuan, H.Z., Zhang, H., Wan, S. and Choo, K.K.R., 2019. Adaptive fusion and category-level dictionary learning model for multi-view human action recognition. IEEE Internet of Things Journal.
 \bibitem{Wang:Wan}
 Gao, Z., Wang, D.Y., Wan, S.H., Zhang, H. and Wang, Y.L., 2019. Cognitive-inspired class-statistic matching with triple-constrain for camera free 3D object retrieval. Future Generation Computer Systems, 94, pp.641-653.
 \bibitem{Kotzias}
 From Group to Individual Labels using Deep Features', Kotzias et. al,. KDD 2015 
 \bibitem{Simeon}
 Simeon, Mondelle and Hilderman, Robert: Categorical Proportional Difference: A Feature Selection Method for Text Categorization, In: Proceedings of the 7th Australasian Data Mining Conference - Vol 87, AusDM '08, pp 201-208, (2008)
 \bibitem{Galavotti}
 Galavotti L., Sebastiani F., Simi M. (2000) Experiments on the Use of Feature Selection and Negative Evidence in Automated Text Categorization. In: Borbinha J., Baker T. (eds) Research and Advanced Technology for Digital Libraries. ECDL 2000. Lecture Notes in Computer Science, vol 1923. Springer, Berlin, Heidelberg.
 \bibitem{Cai:Song}
 Cai J., Song F. (2008) Maximum Entropy Modeling with Feature Selection for Text Categorization. In: Li H., Liu T., Ma WY., Sakai T., Wong KF., Zhou G. (eds) Information Retrieval Technology. AIRS 2008. Lecture Notes in Computer Science, vol 4993. Springer, Berlin, Heidelberg
 \bibitem{Rehman}
 Rehman, A., Javed, K., Babri, H.A., Saeed, M., 2015. Relative discrimination criterion–A novel feature ranking method for text data. Expert Syst. Appl. 42, 3670–3681.
 \bibitem{Metsis}
 Vangelis Metsis, Ion Androutsopoulos, and Georgios Paliouras. 2006. Spam filtering with naive bayes which naive bayes? In Proceedings of CEAS.
 \bibitem{Pang:Lee}
 Pang, B. and Lee, L., 2005, June. Seeing stars: Exploiting class relationships for sentiment categorization with respect to rating scales. In Proceedings of the 43rd annual meeting on association for computational linguistics (pp. 115-124). Association for Computational Linguistics.
 \bibitem{Socher:Alex}
 Socher, R., Perelygin, A., Wu, J., Chuang, J., Manning, C.D., Ng, A. and Potts, C., 2013, October. Recursive deep models for semantic compositionality over a sentiment treebank. In Proceedings of the 2013 conference on empirical methods in natural language processing (pp. 1631-1642).
 \bibitem{Joulin}
 Joulin, A.; Grave, E.; Bojanowski, P.; and Mikolov, T. 2017. Bag of tricks for efficient text classification. In EACL, 427–431. Association for Computational Linguistics.
 \bibitem{Le:Mikolov}
 Le, Q. and Mikolov, T., 2014, January. Distributed representations of sentences and documents. In International conference on machine learning (pp. 1188-1196).
 \bibitem{Xiao}
 Xiao, L., Zhang, H., Chen, W., Wang, Y. and Jin, Y., 2018, July. Transformable Convolutional Neural Network for Text Classification. In IJCAI (pp. 4496-4502).
 \bibitem{Liu:Qiu}
 Liu, P., Qiu, X. and Huang, X., 2016. Recurrent neural network for text classification with multi-task learning. arXiv preprint arXiv:1605.05101.
 \bibitem{Blitzer}
 Blitzer, J., Dredze, M. and Pereira, F., 2007, June. Biographies, bollywood, boom-boxes and blenders: Domain adaptation for sentiment classification. In Proceedings of the 45th annual meeting of the association of computational linguistics (pp. 440-447).
\end{thebibliography}
\end{document}